# Understanding (dis)similarity measures

Lluís A. Belanche
School of Computer Science
Technical University of Catalonia
Jordi Girona, 1-3. 08034 Barcelona, Catalonia, Spain
`belanche@lsi.upc.edu`

**Abstract**

Intuitively, the concept of similarity is the notion to measure an inexact matching between two entities of the same reference set. The notions of similarity and its close relative dissimilarity are widely used in many fields of Artificial Intelligence. Yet they have many different and often partial definitions or properties, usually restricted to one field of application and thus incompatible with other uses. This paper contributes to the design and understanding of similarity and dissimilarity measures for Artificial Intelligence. A formal dual definition for each concept is proposed, joined with a set of fundamental properties. The behavior of the properties under several transformations is studied and revealed as an important matter to bear in mind. We also develop several practical examples that work out the proposed approach.

## 1 Introduction

From a psychological point of view, a human being uses the notions of *similarity* and *dissimilarity* for problem solving, inductive reasoning, element categorization, or simply to search for information partially matching specific criteria. The ability to assess similarities between a newly given pattern and already known patterns is a distinctive feature of human thinking.

It is therefore not strange that similarity and its dual concept dissimilarity are a fundamental part of many theories and applications in several fields, within or related to Artificial Intelligence, like Case Based Reasoning [1], Data Mining [2], Information Retrieval [3], Pattern Matching [4] or Neural Networks, as the Radial Basis Function network [5]. Many applications are characterized by the use of metrics for measuring differences between objects. Metric dissimilarities have been deeply studied but they are tied to a particular transitivity expression based on the triangle inequality. Very often metric (distance) functions are used due to our natural understanding of Euclidean spaces. However, not all metrics are Euclidean and many interesting dissimilarities are non-metric.

In a general sense, similarity and dissimilarity express a dual comparison between two elements. We argue that every property of a similarity should have a correspondence with one property of a dissimilarity and vice versa. This duality is commonly ignored, as well as some annoying properties (e.g. transitivity) and there are few general studies about how transformations of a similarity or dissimilarity can alter their properties. To worsen matters, some properties that would look natural or fundamental –like symmetry or transitivity– are still under discussion (see e.g. [6], [7], [8]). In summary, the lack of a basic agreed-upon theory sometimes leads to incompatible definitions or results focused on an specific kind of similarities or dissimilarities.

The present work intends to make a further effort in the unification of both concepts (see, for example, [9]), in two basic ways. First, with a basic but fully operational definition of similarity and dissimilarity and a set of fundamental properties and transformations. And second, with a study of how these transformations change the properties of the similarities and dissimilarities.

## 2 Preliminaries

Let $X$ be a non-empty set where an equality relation is defined. In a general sense, similarity and dissimilarity express the degree of coincidence or divergence between two elements of a reference set. Therefore, it is reasonable to treat them as functions since the objective is to measure or calculate this value between any two elements of the set.

**Definition 1.** *A* similarity measure *is an upper bounded, exhaustive and total function* $s : X \times X \to I_s \subset \mathbb{R}$ *with* $|I_s| > 1$ *(therefore $I_s$ is upper bounded and* $\sup I_s$ *exists).*

**Definition 2.** *A* dissimilarity measure *is a lower bounded, exhaustive and total function* $d : X \times X \to I_d \subset \mathbb{R}$ *with* $|I_d| > 1$ *(therefore $I_d$ is lower bounded and* $\inf I_d$ *exists).*

Define now $s_{max} = \sup I_s$ and $d_{min} = \inf I_d$. Without loss of generality, we can take $s_{max} \geq 0$ and $d_{min} \geq 0$. In any other case, a non-negative maximum or minimum can be obtained applying a simple transformation (e.g. $s + |s_{max}|$). The following are useful properties for these functions to fulfill. For conciseness, we introduce them for both kinds of functions at the same time.

**Reflexivity**: $s(x, x) = s_{max}$ (implying $\sup I_s \in I_s$) and $d(x, x) = d_{min}$ (implying $\inf I_d \in I_d$).
**Strong Reflexivity**: $s(x, y) = s_{max} \Leftrightarrow x = y$ and $d(x, y) = d_{min} \Leftrightarrow x = y$.
**Symmetry**: $s(x, y) = s(y, x)$ and $d(x, y) = d(y, x)$.
**Boundedness**: A similarity $s$ is *lower* bounded when $\exists a \in R$ such that $s(x, y) \geq a$, for all $x, y \in X$ (this is equivalent to ask that $\inf I_s$ exists). Conversely, a dissimilarity $d$ is *upper* bounded when $\exists a \in R$ such that $d(x, y) \leq a$, for all $x, y \in X$ (this is equivalent to ask that $\sup I_d$ exists). Given that $|I_s| > 1$ and $|I_d| > 1$, both $\inf I_s \neq \sup I_s$ and $\inf I_d \neq \sup I_d$ hold true.
**Closedness**: Given a lower bounded function $s$, define now $s_{min} = \inf I_s$. The property asks for the existence of $x, y \in X$ such that $s(x, y) = s_{min}$ (equivalent to asking that $\inf I_s \in I_s$). Given an upper bounded function $d$, define $d_{max} = \sup I_d$. The property asks for the existence of $x, y \in X$ such that $d(x, y) = d_{max}$ (equivalent to asking that $\sup I_d \in I_d$).

**Complementarity**: Consider now a function $C : X \to 2^X$. A lower closed similarity $s$ defined in $X$ has *complement function* $C(x) = \{x' \in X / s(x, x') = s_{min}\}$, if $\forall x, x' \in X, |C(x)| = |C(x')| \neq 0$. An upper closed dissimilarity $d$ defined in $X$ has complement function $C$, where $C(x) = \{x' \in X / d(x, x') = d_{max}\}$, if $\forall x, x' \in X, |C(x)| = |C(x')| \neq 0$. In case $s$ or $d$ are reflexive, necessarily $x \notin C(x)$. Each of the elements in $C(x)$ will be called a *complement* of $x$. Moreover, $s$ or $d$ have *unitary* complement when $\forall x \in X, |C(x)| = 1$. In this case, $\forall x \in X$:

For similarities: $\exists y' / s(x, y') = s_{max} \iff \exists y' / y' \in C(y), \forall y \in C(x)$

For dissimilarities: $\exists y' / d(x, y') = d_{min} \iff \exists y' / y' \in C(y), \forall y \in C(x)$

Let us define a *transitivity* operator in order to introduce the transitivity property in similarity and dissimilarity functions.

**Definition 3.** *(Transitivity operator). Let $I$ be a non-empty subset of $\mathbb{R}$, and let $e$ be a fixed element of $I$. A transitivity operator is a function $\tau : I \times I \to I$ satisfying, for all $x, y, z \in I$:*

1. $\tau(x, e) = x$ *(null element)*

2. $y \leq z \Rightarrow \tau(x, y) \leq \tau(x, z)$ *(non-decreasing monotonicity)*

3. $\tau(x, y) = \tau(y, x)$ *(symmetry)*

4. $\tau(x, \tau(y, z)) = \tau(\tau(x, y), z)$ *(associativity)*

There are two groups of transitivity operators: those for similarity functions, for which $e = \sup I = s_{max}$ (and then $I$ is $I_s$) and those for dissimilarity functions, for which $e = \inf I = d_{min}$ ($I$ is $I_d$). It should be noted that this definition reduces to uninorms [10] when $I = [0, 1]$.

**Transitivity**: A similarity $s$ defined in $X$ is called $\tau_s$-transitive if there is a transitivity operator $\tau_s$ such that the following inequality holds:

$$s(x, y) \geq \tau_s(s(x, z), s(z, y)) \; \forall x, y, z \in X$$

A dissimilarity $d$ defined in $X$ is called $\tau_d$-transitive if there is a transitivity operator $\tau_d$ such that the following inequality holds:

$$d(x, y) \leq \tau_d(d(x, z), d(z, y)) \; \forall x, y, z \in X$$

A similarity or dissimilarity in $X$ may be required simply to satisfy strong reflexivity and symmetry. It is not difficult to show that strong reflexivity alone implies a basic form of transitivity [11]. We call $\Sigma(X)$ the set of all similarity functions and $\Delta(X)$ the set of all dissimilarity functions defined over elements of $X$.

## 3 Equivalence

In this section we tackle the problem of obtaining *equivalent* similarities or dissimilarities, and to transform a similarity function onto a dissimilarity function or vice versa, which will naturally lead to the concept of *duality*.

## 3.1 Equivalence functions

Consider the set of all ordered pairs of elements of $X$ and denote it $X \times X$. Every $s \in \Sigma(X)$ induces a preorder relation in $X \times X$. This preorder is defined as "to belong to a class of equivalence with less or equal similarity value". Formally, given $X$ and $s \in \Sigma(X)$, we consider the preorder $\preceq$ given by

$$(x,y) \preceq (x',y') \iff s(x,y) \leq s(x',y'), \forall (x,y), (x',y') \in X \times X$$

Analogously, every $d \in \Delta(X)$ induces the preorder "to belong to a class of equivalence with less or equal dissimilarity value". Recall that $(x,y) \preceq (w,z)$ and $(w,z) \preceq (x,y)$ does *not* imply $x = w$ and $y = z$.

**Definition 4.** *(Equivalence). Two similarities (or two dissimilarities) defined in the same reference set $X$ are* equivalent *if they induce the same preorder.*

Note that the equivalence between similarities or between dissimilarities is an equivalence relation. The properties of similarities and dissimilarities are kept under equivalence, including transitivity. The exception is the boundedness property which will depend on the chosen equivalence function. Only the *monotonically increasing* and *invertible* functions keep the induced preorder.

**Definition 5.** *(Equivalence function). Let $s$ be a similarity and $d$ a dissimilarity. An equivalence function is a monotonically increasing and invertible function $\breve{f}$ such that $\breve{f} \circ s$ is a similarity equivalent to $s$. Analogously, $\breve{f} \circ d$ is a dissimilarity equivalent to $d$.*

**Theorem 1.** *Let $s_1$ be a transitive similarity and $d_1$ a transitive dissimilarity. Denote by $\tau_{s_1}$ and $\tau_{d_1}$ their respective transitivity operators. Let $\breve{f}$ be an equivalence function. Then:*

1. *The equivalent similarity $s_2 = \breve{f} \circ s_1$ is $\tau_{s_2}$-transitive, where*

   $\tau_{s_2}(a,b) = \breve{f}(\tau_{s_1}(\breve{f}^{-1}(a), \breve{f}^{-1}(b))) \ \forall a,b \in I_{s_2}$

2. *The equivalent dissimilarity $d_2 = \breve{f} \circ d_1$ is $\tau_{d_2}$-transitive, where*

   $\tau_{d_2}(a,b) = \breve{f}(\tau_{d_1}(\breve{f}^{-1}(a), \breve{f}^{-1}(b))) \ \forall a,b \in I_{d_2}$

   *Proof.* Consider only the similarity case, in which $\breve{f}: I_{s_1} \to I_{s_2}$. Using the transitivity of $s_1$ we know that, for all $x,y,z \in X$, $s_1(x,y) \geq \tau_{s_1}(s_1(x,z), s_1(z,y))$.
   Applying $\breve{f}$ to this inequality we get
   $(\breve{f} \circ s_1)(x,y) \geq (\breve{f} \circ \tau_{s_1})(s_1(x,z), s_1(z,y))$.
   Using $\breve{f}^{-1} \circ s_2 = s_1$, we get
   $s_2(x,y) \geq (\breve{f} \circ \tau_{s_1}) \left( (\breve{f}^{-1} \circ s_2)(x,z), (\breve{f}^{-1} \circ s_2)(z,y) \right)$.
   Defining $\tau_{s_2}$ as is defined in the Theorem we get the required transitivity expression $s_2(x,y) \geq \tau_{s_2}(s_2(x,z), s_2(z,y))$.

Therefore, any composition of an equivalence function and a similarity (or dissimilarity) function is another similarity (or dissimilarity) function, which is also equivalent.

## 3.2 Transformation functions

Equivalence functions allow us to get new similarities from other similarities or new dissimilarities from other dissimilarities, but not to switch between the former and the latter. Denote by $\Sigma^*(X)$ the set of similarities defined in $X$ with codomain on [0,1] and by $\Delta^*(X)$ the set of such dissimilarities. As we shall see, using appropriate equivalence functions $\breve{f}^*$, we have a way to get equivalent similarities (resp. dissimilarities) on $\Sigma^*(X)$ (resp. $\Delta^*(X)$) using similarities or dissimilarities in $\Sigma(X)$ (resp. $\Delta(X)$) and vice versa. In consequence, defining properties in $\Sigma(X)$ or $\Delta(X)$ is tantamount to defining them in $\Sigma^*(X)$ or $\Delta^*(X)$, respectively.

**Definition 6.** *A $[0,1]$-transformation function $\hat{n}$ is a decreasing bijection on [0,1] (implying that $\hat{n}(0) = 1, \hat{n}(1) = 0$, continuity and the existence of an inverse). A transformation function $\hat{n}$ is involutive if $\hat{n}^{-1} = \hat{n}$.*

This definition is restricted to (resp. dissimilarities) on $\Sigma^*(X)$ (resp. $\Delta^*(X)$). Using that both $\breve{f}^*$ and $\hat{n}$ are bijections, a general transformation function between elements of $\Sigma(X)$ (resp. $\Delta(X)$) is the composition of two or more functions in the following way:

**Definition 7.** *A transformation function $\hat{f}$ is the composition of two equivalence functions and a $[0,1]$-transformation function:*
$$\hat{f} = \breve{f}^*_1 \circ \hat{n} \circ \breve{f}^{*^{-1}}_2,$$
*where $\hat{n}$ is a transformation function on [0,1], $\breve{f}^*_1$ obtains equivalent similarities (resp. dissimilarities) in $\Sigma(X)$ (resp. $\Delta(X)$) and $\breve{f}^*_2$ obtains equivalent similarities (resp. dissimilarities) in $\Sigma^*(X)$ (resp. $\Delta^*(X)$).*

## 4 Duality

As it has been shown along this work, similarity and dissimilarity are two interrelated concepts. In fuzzy theory, t-norms and t-conorms are dual with respect to the fuzzy complement [12]. In the same sense, all similarity and dissimilarity functions are dual with respect to some transformation function.

**Definition 8.** *(Duality). Consider $s \in \Sigma(X), d \in \Delta(X)$ and a transformation function $\hat{f} : I_s \to I_d$. We say that $s$ and $d$ are dual by $\hat{f}$ if $d = \hat{f} \circ s$ or, equivalently, if $s = \hat{f}^{-1} \circ d$. This relationship is written as a triple $\prec s, d, \hat{f} \succ$.*

**Theorem 2.** *Given a dual triple $\prec s, d, \hat{f} \succ$,*

1. *$d$ is strongly reflexive if and only if $s$ is strongly reflexive.*

2. *$d$ is closed if and only if $s$ is closed.*

3. *$d$ has (unitary) complement if and only if $s$ has (unitary) complement.*

4. *$d$ is $\tau_d$-transitive only if $s$ is $\tau_s$-transitive, where*
   $$\tau_d(x, y) = \hat{f}(\tau_s(\hat{f}^{-1}(x), \hat{f}^{-1}(y))) \; \forall x, y \in I_d$$

*Proof.* Take $s \in \Sigma(X)$ and make $d = \hat{f} \circ s$.

1. For all $x, y \in X$ such that $x \neq y$, we have $s(x,y) \neq s_{max}$; hence, applying $\hat{f}$, we obtain $d(x,y) \neq d_{min}$.

2. Symmetry is immediate.

3. For all $x, y \in X$, we have $s(x,y) \geq s_{min}$. Suppose $s$ is closed. Since $\hat{f}$ is strictly monotonic and decreasing, $s(x,y) > s_{min} \Leftrightarrow (\hat{f} \circ s)(x,y) < \hat{f}(s_{min})$. Then $s$ is closed because there exist $x, y \in X$ such that $s(x,y) = s_{min}$, only true if $(\hat{f} \circ s)(x,y) = \hat{f}(s_{min})$ (i.e. if $d$ is closed).

4. For all $x, x' \in X$ such that $x' \in C(x)$, we have $s(x,x') = s_{min}$; applying $\hat{f}$, we have $(\hat{f} \circ s)(x, x') = \hat{f}(s_{min})$; that is, $d(x, x') = d_{max}$. Therefore, complementarity is kept.

5. For transitivity, see [12], Theorem 3.20, page 84.

Thanks to this explicit duality relation, properties on similarities are immediately translated to dissimilarities, or viceversa. A general view of all the functions and sets appeared so far is represented in Fig. 4.1.

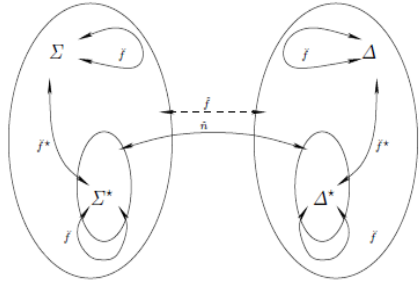

Figure 4.1: Graphical representation of equivalence ($\check{f}$) and transformation ($\hat{f}$) functions from and within $\Sigma(X)$ and $\Delta(X)$.

## 5 Application examples

In this section we develop some simple application examples for the sake of illustration.

**Example 1.** *Consider the dissimilarities in $\Sigma([0,1])$ given by*

$$d_1(x,y) = |x - y|, \quad d_2(x,y) = min(x,y).$$

*Their respective transitivity operators are $\tau_{d_1}(a,b) = min(1, a+b)$ and $\tau_{d_2}(a,b) = min(a,b)$. Consider the family of transformation functions: $\hat{f}(z) = (1-z)^{1/\alpha}$, with $\alpha \neq 0$. The corresponding dual similarities are:*

$$s_1(x,y) = (1 - |x-y|)^{1/\alpha}, \quad s_2(x,y) = \max((1-x)^{1/\alpha}, (1-y)^{1/\alpha}).$$

Using Theorem 2, the corresponding transitivity operators are $\tau_{s_1}(a,b) = \max(a^\alpha + b^\alpha - 1, 0)^{1/\alpha}$ and $\tau_{s_2}(a,b) = max(a,b)$. Therefore, two dual triples are formed: $\prec s_1, d_1, \hat{f} \succ$ and $\prec s_2, d_2, \hat{f} \succ$. Note that $\tau_{s_1}$ corresponds to a well-known family of t-norms, whereas $\tau_{s_2}$ is the max norm. When $\alpha = 1$, the transitivity of $s_1$ is the Lukasiewicz t-norm [13].

**Example 2.** *Consider the similarity defined in $\Sigma(\mathbb{Z})$ given by $s(x,y) = 1 - \frac{|x-y|}{|x-y|+1}$. In this case the set $I_s$ is the set of all rational numbers in $(0,1]$, $\sup I_s = 1$ and $\inf I_s = 0$. This function satisfies strong reflexivity and symmetry. Moreover, it is lower bounded (with $s_{min} = 0$), although it is not lower closed. For this reason, it does* not *have a complement function.*

What transitivity do we have here? We know that $|x-y|$ is a metric. Consider now the transformations $\hat{n}_k(z) = z/(z+k)$, for $k > 0$. Since $\hat{n}_k$ is subadditive, $\hat{n}_k(|x-y|)$ is also a metric dissimilarity. Therefore,

$$\frac{|x-y|}{|x-y|+1} \leq \frac{|x-z|}{|x-z|+1} + \frac{|z-y|}{|z-y|+1}$$

If we apply now the transformation $\hat{n}(z) = 1 - z$, we obtain the original expression for the similarity $s$. Using Theorem 2, the transitivity finally changes to $s(x,y) = \max\{s(x,z) + s(z,y) - 1, 0\}$.

**Example 3.** *Consider the function $d(x,y) = e^{|x-y|} - 1$. This is a strong reflexive and symmetric dissimilarity in $\Delta(\mathbb{R})$ with codomain $I_d = [0, +\infty)$. Therefore, it is an unbounded dissimilarity with $d_{min} = 0$. This measure can be expressed as the composition of $\breve{f}(z) = e^z - 1$ and $d'(x,y) = |x-y|$. Thus, it is $\tau_d$-transitive with $\tau_d(a,b) = ab + a + b$. Consequently,*

$$d(x,y) \leq d(x,z) + d(z,y) + d(x,z) \cdot d(z,y), \ \forall x,y,z \in \mathbb{R}$$

To see this, use that $d'$ is $d'$-transitive with $\tau_{d'}(a,b) = a + b$ and apply Theorem 1:

$$\tau_d(a,b) = \breve{f}(\tau_{d'}[\breve{f}^{-1}(a), \breve{f}^{-1}(b)]) = e^{\ln(1+a) + ln(1+b)} - 1 = (1+a)(1+b) - 1 = ab + a + b$$

Consider now the equivalence function $\breve{f} : [0, \infty) \to [0, \infty)$ given by $\breve{f}(z) = ln(z+1)$ and apply it to the previously defined dissimilarity $d$. The result is the *equivalent* dissimilarity $d'(a,b) = |x-y|$, the standard metric in $\mathbb{R}$, transitive with $\tau_{d'}(a,b) = a + b$ (this is the transitivity leading to the triangular inequality for metrics). The important point is that $d'$ is *also* $\tau_d$-transitive, since $a + b \leq a + b + ab$ when $a,b \in [0, \infty)$. This is due to a *gradation* in the restrictiveness of transitivity operators [12]. In this case, $d'$ is *more restrictive* than $d$ and therefore, transitivity with the former operator implies transitivity with the latter, but not inversely.

If we apply now $\breve{f}'(z) = z^2$ to $d'$ what we get is an equivalent dissimilarity $d''(x,y) = (x-y)^2$, again strongly reflexive, symmetric and $d''$-transitive, where $\tau_{d''}(a,b) = \sqrt{a^2 + b^2}$. In this case, $d''$ is more restrictive than both $d'$ and $d$.

Similarity and dissimilarity unify preservation of transitivity using equivalence functions. This fact can be used, for example, to get a metric dissimilarity from a non-metric one. In the following example we compare the structure of two trees with a non-metric dissimilarity. Upon application of an equivalence function we get an *equivalent* and *metric* dissimilarity function.

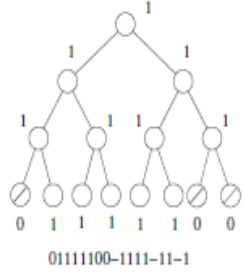

Figure 5.1: A simple coding of binary trees. The reason for going bottom-up is to have the less significative digits close to the root of the tree. The choice of making the left nodes more significant than the right ones is arbitrary. The symbol ⊘ represents the empty tree.

**Example 4.** *Consider a dissimilarity function between two binary trees. It does not measure differences between nodes but the structure of the tree. Consider a simple tree coding function D that assigns a unique value for each tree. This value is first coded as a binary number of length $2^h - 1$, being h the height of the tree. The reading of the code as a natural number is the tree code. The binary number is computed such that the most significant bit corresponds to the leftmost and bottommost tree node (Fig. 5.1). Note that D is not a bijection, since there are numbers that do not code a valid binary tree.*

*Consider now the following dissimilarity function, where A and B are binary trees. The symbol ⊘ represents the empty tree with value 0.*

$$d(A,B) = \begin{cases} \max\left(\frac{D(A)}{D(B)}, \frac{D(B)}{D(A)}\right) & \text{if } A \neq \oslash \text{ and } B \neq \oslash \\ 1 & \text{if } A = \oslash \text{ and } B = \oslash \\ D(A) & \text{if } A \neq \oslash \text{ and } B = \oslash \\ D(B) & \text{if } A = \oslash \text{ and } B \neq \oslash \end{cases}$$

This is a strong reflexive, symmetric, unbounded dissimilarity with $I_d = [1, \infty)$ with $d_{min} = 1$. If we impose a limit $H$ to the height of the trees, then $d$ is also upper bounded and closed, $d_{max} = \sum_{i=0}^{2^H - 1} 2^i$. It is also transitive with the *product* operator, which is a transitivity operator valid for dissimilarities defined in $[1, \infty)$; in other words, for any three trees $A, B, C$, $d(A, B) \leq d(A, C) \cdot d(C, B)$.

*Proof.* If neither of $A, B$ or $C$ are the empty tree, substituting in the previous expression and operating with max and the product we get:

$$\max\left(\frac{D(A)}{D(B)}, \frac{D(B)}{D(A)}\right) \leq \max\left(\frac{D(A)}{D(B)}, \frac{D(C)^2}{D(A)D(B)}, \frac{D(A)D(B)}{D(C)^2}, \frac{D(B)}{D(A)}\right)$$

which is trivially true. Now, if $A = \oslash$, then the inequality reduces to $D(B) \leq \max\left(D(B), \frac{D(C)^2}{D(A)D(B)}\right)$. The cases $B = \oslash$ or $C = \oslash$ can be treated analogously.

If we apply now the equivalence function $\breve{f}(z) = \log z$ to $d$ we shall receive a dissimilarity $d' = \breve{f} \circ d$, where the properties of $d$ are kept in $d'$. However, the transitivity operator is changed using Theorem 1, to $\tau_{d'}(a, b) = a + b$. In other words, we obtain a metric dissimilarity over trees fully equivalent to the initial choice of $d$.

# 6 Conclusions

The main goal of this paper has not been to set up a standard definition of similarity and dissimilarity, but to establish some operative grounds on the definition of these widely used concepts. The data practitioner can take (or leave) the proposed properties as a guide. We have studied some fundamental transformations in order to keep these chosen basic properties. In particular, we have concentrated on transitivity and its preservation. However, a deeper study has to be done about the effects of transformations, specially in transitivity (e.g. which transformations do keep the triangle inequality) and more complex matters, like aggregation of different measures into a global one. Due to the many fields of application these concepts are involved with, the study of their properties can lead to better understanding of similarity and dissimilarity measures in many areas.

# References

[1] H. Osborne, D. Bridge: Models of similarity for case-based reasoning. In: Interdisciplinary Workshop on Similarity and Categorisation, pp. 173–179 (1997)

[2] T. Li, Z. Shenghuo, M. Ogihara: A new distributed data mining model based on similarity. In: ACM SAC Data Mining Track, Florida, USA (2003)

[3] R. Baeza-Yates, B. Ribeiro-Neto: Modern information Retrieval, ACM Press, New York (1999)

[4] R.C. Veltkamp, M. Hagedoorn: Shape similarity measures, properties and constructions. In Advances in Visual Information Systems, VISUAL 2000. LNCS, vol. 1929, Springer, Heidelberg (2000)

[5] S. Haykin: Neural Networks: A Comprehensive Foundation. Prentice Hall (1998)

[6] A. Tversky: Features of similarity. Psycological Review 84(4), 327–352 (1977)

[7] M. DeCock, E. Kerre: On (un)suitable relations to model approximate equality. Fuzzy Sets And Systems 133, 137–153 (2003)

[8] S. Santini, R. Jain: Similarity measures. IEEE Transactions on Pattern Analysis and Machine Intelligence (1999).

[9] H. Bock, E. Diday: Analysis of symbolic data. Exploratory Methods for Extracting Statistical Information from Complex Data. Springer (1999)

[10] E. P. Klement: Some mathematical aspects on fuzzy sets: Triangular norms, fuzzy logics, generalized measures. Fuzzy Sets And Systems 90, 133–140 (1997)

[11] J. Orozco: Similarity and dissimilarity concepts in machine learning. Technical Report LSI-04-9-R, Universitat Politcnica de Catalunya, Barcelona, Spain (2004)

[12] G. Klir, B. Yuan: Fuzzy Sets and Fuzzy Logic: Theory and Applications. Pearson Education (1995)

[13] B. Schweizer, A. Sklar: Probabilistic Metric Spaces, North-Holland, Amsterdam (1983)